\documentclass[11pt,letterpaper]{article}

\pdfoutput=1

\usepackage{amsmath}
\usepackage{enumitem}
\usepackage{ccg}
\usepackage{graphicx}
\usepackage{xspace}
\usepackage{multirow}
\usepackage{url}
\usepackage{algorithm} 
\usepackage{algorithmicx}
\usepackage[noend]{algpseudocode}
\usepackage{cleveref}

\usepackage{subcaption}
\usepackage{tikz,pgfplots}
    \usepgfplotslibrary{groupplots}

\usetikzlibrary{positioning}
\usetikzlibrary{patterns}
\usetikzlibrary{trees}

\usetikzlibrary{plotmarks}
\pgfplotsset{
  tick label style = {font=\scriptsize},
  every axis label = {font=\footnotesize},
  legend style = {font=\footnotesize},
  label style = {font=\footnotesize},
}

\newcommand{\ignore}[1]{}
\newcommand{\astar}[0] {A$^{*}$\xspace}
\newcommand*\Funct[2]{\textsc{#1}(#2)}
\newcommand*\Let[2]{\State #1 $\gets$ #2}
\newcommand{\agenda}[0] {\mathcal{A}\xspace}
\newcommand{\forest}[0] {\mathcal{F}\xspace}
\newcommand{\violation}[0] {\mathcal{V}\xspace}
\newcommand{\argmax}{\operatornamewithlimits{argmax}}

\newcommand{\unknownrule}[1]
{ \mc{#1}{\hrulefill_{?}} }

\definecolor{forest}{RGB}{68, 110, 182}
\definecolor{frontier}{RGB}{246, 180, 32}
\definecolor{unexplored}{RGB}{240, 240, 240}

\newcommand\forestnode[4] {
\node[text=white](#1) at (#2, #3) {
\fcolorbox{black}{forest}{
\textcolor{white}{
\hspace{-0.5cm}
#4
\hspace{-0.4cm}
}
}
};
}
\newcommand\frontiernode[4] {
\node(#1) at (#2, #3) {
\fcolorbox{black}{frontier}{
\hspace{-0.5cm}
#4
\hspace{-0.4cm}
}
};
}
\newcommand\unexplorednode[4] {
\node(#1) at (#2, #3) {
\fcolorbox{black}{unexplored}{
\hspace{-0.5cm}
#4
\hspace{-0.4cm}
}
};
}
\newcommand\forestedge[4] {
\draw(#1)[->, line width=2pt] to[out=#2,in=#3] (#4);
}
\newcommand\frontieredge[4] {
\draw(#1)[->, line width=0.7pt] to[out=#2,in=#3] (#4);
}
\newcommand\unexplorededge[4] {
\draw(#1)[->, dashed] to[out=#2,in=#3] (#4);
}

\usepackage{times}
\usepackage{latexsym}
\usepackage{emnlp2016}

\emnlpfinalcopy

\def\emnlppaperid{910}

\title{Global Neural CCG Parsing with Optimality Guarantees}

\author{Kenton Lee \qquad Mike Lewis \qquad Luke Zettlemoyer \\
Computer Science \& Engineering\\
University of Washington\\
Seattle, WA 98195 \\
\texttt{\{kentonl,mlewis,lsz\}@cs.washington.edu}\\
}

\date{}

\begin{document}

\maketitle

\begin{abstract}
We introduce the first global recursive neural parsing model with optimality guarantees during decoding. To support global features, we give up dynamic programs and instead search directly in the space of all possible subtrees. 
Although this space is exponentially large in the sentence length, we show it is possible to learn an efficient A* parser. We augment existing parsing models, which have informative bounds on the outside score, with a global model that has loose bounds but only needs to model non-local phenomena.
The global model is trained with a novel objective that encourages the parser to search both efficiently and accurately.
The approach is applied to CCG parsing, improving state-of-the-art accuracy by 0.4~F1. The parser finds the optimal parse for 99.9\% of held-out sentences, exploring on average only 190 subtrees.

%The model is trained with respect to an A* search agenda, allowing it to learn to efficiently search through exponentially many structures, despite no dynamic programming. 
%The approach outperforms both factored and greedy methods, in both efficiency and accuracy.

\end{abstract}

\section{Introduction}

Recursive neural models perform well for many structured prediction problems, in part due to their ability to learn representations that depend globally on all parts of the output structures. 
%Recursive neural models are well suited for structured prediction, as they can embed and score entire substructures. 
However, global models of this sort are incompatible with existing exact inference algorithms, since they do not decompose over substructures in a way that allows effective dynamic programming.
Existing work has therefore used greedy inference techniques such as beam search~\cite{vinyals:2014,dyer:2015} or reranking~\cite{socher:2013}. %, and it is unclear how much these approximations hurt performance in practice. 
We introduce the first global recursive neural parsing approach with optimality guarantees for decoding and use it to build a state-of-the-art CCG parser.

\begin{figure*}[ht!]
\begin{subfigure}[b]{0.45\linewidth}
\frame{
\scalebox{0.49}{
\begin{tikzpicture}
\clip(-7,-6) rectangle (9, 11);
\forestnode{start}{0}{0}{
\hspace{0.3cm}
$\emptyset$
\hspace{0.2cm}
}
\frontiernode{fruit2}{-5.5}{-1.5}{
\deriv{1}{
\mc{1}{\mbox{Fruit}}\\
\uline{1}\\
\mc{1}{NP} \\
}
};
\frontiernode{flies2}{-1}{-1.5}{
\deriv{1}{
\mc{1}{\mbox{flies}}\\
\uline{1}\\
\mc{1}{S \bs NP} \\
}
};
\frontiernode{like2}{3}{-1.5}{
\deriv{1}{
\mc{1}{\mbox{like}}\\
\uline{1}\\
\mc{1}{(S \bs S) /NP} \\
}
};
\frontiernode{flies3}{-3.8}{2}{
\deriv{1}{
\mc{1}{\mbox{flies}}\\
\uline{1}\\
\mc{1}{NP \bs NP} \\
}
};
\forestnode{fruit}{-1.6}{2}{
\deriv{2}{
\mc{1}{\mbox{Fruit}}\\
\uline{1}\\
\mc{1}{NP/NP}\\
}
};
\forestnode{flies}{0.25}{2}{
\deriv{2}{
\mc{1}{\mbox{flies}}\\
\uline{1}\\
\mc{1}{NP}\\
}
};
\forestnode{like}{2.5}{2}{
\deriv{2}{
\mc{1}{\mbox{like}}\\
\uline{1}\\
\mc{1}{(S \bs NP) /NP}\\
}
};
\forestnode{bananas}{5}{2}{
\deriv{2}{
\mc{1}{\mbox{bananas}}\\
\uline{1}\\
\mc{1}{NP}\\
}
};
\forestnode{fruit-flies}{-2.5}{5}{
\deriv{2}{
\mc{1}{\mbox{Fruit}} & \mc{1}{\mbox{flies}}\\
 \unknownrule{2}\\
 \mc{2}{NP}\\
}
};
\forestnode{like-bananas}{2.5}{5}{
\deriv{2}{
\mc{1}{\mbox{like}} & \mc{1}{\mbox{bananas}}\\
 \unknownrule{2}\\
 \mc{2}{S\bs NP}\\
}
};
\unexplorednode{like-bananas2}{6}{-4}{
\deriv{2}{
\mc{1}{\mbox{like}} & \mc{1}{\mbox{bananas}}\\
 \unknownrule{2}\\
 \mc{2}{S\bs S}\\
}
};
\unexplorednode{fruit-flies2}{-5.5}{-4}{
\deriv{2}{
\mc{1}{\mbox{Fruit}} & \mc{1}{\mbox{flies}}\\
 \unknownrule{2}\\
 \mc{2}{S}\\
}
};

{
\fboxrule=2pt%border thickness
\forestnode{fruit-flies-like-bananas}{0.2}{7.5}{
\deriv{4}{
\mc{1}{\mbox{Fruit}} &\mc{1}{\mbox{flies}} & \mc{1}{\mbox{like}} & \mc{1}{\mbox{bananas}}\\
\unknownrule{4} \\
\mc{4}{S}\\
}
}};
\forestedge{start}{150}{-90}{fruit};
\forestedge{start}{90}{-90}{flies};
\forestedge{start}{30}{-90}{like};
\forestedge{start}{15}{-90}{bananas};
\forestedge{like}{90}{-60}{like-bananas};
\forestedge{bananas}{90}{-60}{like-bananas};
\forestedge{fruit}{90}{-60}{fruit-flies};
\forestedge{flies}{90}{-60}{fruit-flies};
\forestedge{fruit-flies}{60}{-90}{fruit-flies-like-bananas};
\forestedge{like-bananas}{120}{-90}{fruit-flies-like-bananas};

\frontieredge{start}{-165}{90}{fruit2};
\frontieredge{start}{-90}{45}{flies2};
\frontieredge{start}{-15}{90}{like2};
\frontieredge{start}{160}{-90}{flies3};
\unexplorededge{fruit2}{95}{-150}{fruit-flies};
\unexplorededge{flies3}{90}{-150}{fruit-flies};

\unexplorededge{like2}{-90}{90}{like-bananas2};
\unexplorededge{bananas}{-30}{90}{like-bananas2};

\unexplorededge{fruit2}{-90}{45}{fruit-flies2};
\unexplorededge{flies2}{-90}{45}{fruit-flies2};

\unexplorededge{fruit-flies2}{125}{90}{fruit-flies-like-bananas};
\unexplorededge{like-bananas2}{0}{90}{fruit-flies-like-bananas};
\draw (5, 7.4) rectangle (8.9, 10.65);

\coordinate (explored-dummy) at (5.25, 10);
\forestnode{explored-node}{7.5}{10}{\hspace{0.4cm}\;\;explored\;\;\hspace{0.25cm}};
\forestedge{explored-dummy}{0}{180}{explored-node};

\coordinate (agenda-dummy) at (5.25, 9);
\frontiernode{agenda-node}{7.5}{9}{\hspace{0.4cm}\;\;\;agenda\;\;\;\hspace{0.25cm}};
\frontieredge{agenda-dummy}{0}{180}{agenda-node};

\coordinate (unexplored-dummy) at (5.25, 8);
\unexplorednode{unexplored-node}{7.5}{8}{\hspace{0.4cm}unexplored\hspace{0.25cm}};
\unexplorededge{unexplored-dummy}{0}{180}{unexplored-node};

\end{tikzpicture}
}
}
\caption{\footnotesize The search space in chart parsing, with one node for each labeling of a span.}
\label{fig:chart-hypergraph}
\end{subfigure}
\hfill
\begin{subfigure}[b]{0.47\linewidth}
\frame{
\scalebox{0.49}{
\begin{tikzpicture}
\clip(-8,-7) rectangle (7.2, 10);
\forestnode{start}{0}{0}{
\hspace{0.3cm}
$\emptyset$
\hspace{0.2cm}
}
\frontiernode{fruit2}{-5.5}{-1.5}{
\deriv{1}{
\mc{1}{\mbox{Fruit}}\\
\uline{1}\\
\mc{1}{NP} \\
}
};
\frontiernode{flies2}{-1}{-1.5}{
\deriv{1}{
\mc{1}{\mbox{flies}}\\
\uline{1}\\
\mc{1}{S \bs NP} \\
}
};
\frontiernode{like2}{3}{-1.5}{
\deriv{1}{
\mc{1}{\mbox{like}}\\
\uline{1}\\
\mc{1}{(S \bs S) /NP} \\
}
};
\frontiernode{flies3}{-4.5}{2}{
\deriv{1}{
\mc{1}{\mbox{flies}}\\
\uline{1}\\
\mc{1}{NP \bs NP} \\
}
};
\forestnode{fruit}{-2}{2}{
\deriv{2}{
\mc{1}{\mbox{Fruit}}\\
\uline{1}\\
\mc{1}{NP/NP}\\
}
};
\forestnode{flies}{0}{2}{
\deriv{2}{
\mc{1}{\mbox{flies}}\\
\uline{1}\\
\mc{1}{NP}\\
}
};
\forestnode{like}{2.5}{2}{
\deriv{2}{
\mc{1}{\mbox{like}}\\
\uline{1}\\
\mc{1}{(S \bs NP) /NP}\\
}
};
\forestnode{bananas}{5}{2}{
\deriv{2}{
\mc{1}{\mbox{bananas}}\\
\uline{1}\\
\mc{1}{NP}\\
}
};
\forestnode{fruit-flies}{-1}{5}{
\deriv{2}{
\mc{1}{\mbox{Fruit}} & \mc{1}{\mbox{flies}}\\
\uline{1} & \uline{1}\\
\mc{1}{NP/NP} & \mc{1}{NP}\\
 \fapply{2}\\
 \mc{2}{NP}\\
}
};
\unexplorednode{fruit-flies3}{-4.5}{5}{
\deriv{2}{
\mc{1}{\mbox{Fruit}} & \mc{1}{\mbox{flies}}\\
\uline{1} & \uline{1}\\
\mc{1}{NP} & \mc{1}{NP \bs NP}\\
 \bapply{2}\\
 \mc{2}{NP}\\
}
};
\forestnode{like-bananas}{3}{5}{
\deriv{2}{
\mc{1}{\mbox{like}} & \mc{1}{\mbox{bananas}}\\
\uline{1} & \uline{1}\\
\mc{1}{(S\bs NP)/NP} & \mc{1}{NP}\\
 \fapply{2}\\
 \mc{2}{S\bs NP}\\
}
};
\unexplorednode{like-bananas2}{5}{-4}{
\deriv{2}{
\mc{1}{\mbox{like}} & \mc{1}{\mbox{bananas}}\\
\uline{1} & \uline{1}\\
\mc{1}{(S\bs S)/NP} & \mc{1}{NP}\\
 \fapply{2}\\
 \mc{2}{S\bs S}\\
}
};
\unexplorednode{fruit-flies2}{1}{-4}{
\deriv{2}{
\mc{1}{\mbox{Fruit}} & \mc{1}{\mbox{flies}}\\
\uline{1} & \uline{1}\\
\mc{1}{NP} & \mc{1}{S \bs NP}\\
 \bapply{2}\\
 \mc{2}{S}\\
}
};
{
\fboxrule=2pt%border thickness
\forestnode{fruit-flies-like-bananas}{3.5}{8.5}{
\deriv{4}{
\mc{1}{\mbox{Fruit}} &\mc{1}{\mbox{flies}} & \mc{1}{\mbox{like}} & \mc{1}{\mbox{bananas}}\\
\uline{1} & \uline{1} & \uline{1} & \uline{1}\\
\mc{1}{NP/NP} & \mc{1}{NP} & \mc{1}{(S\bs NP)/NP} &\mc{1}{NP}\\
\fapply{2} & \fapply{2}\\
 \mc{2}{NP}  & \mc{2}{S\bs NP}\\
\bapply{4}\\
\mc{4}{S}\\
}
}};
{
\fboxrule=2pt%border thickness
\unexplorednode{fruit-flies-like-bananas3}{-4}{8.5}{
\deriv{4}{
\mc{1}{\mbox{Fruit}} &\mc{1}{\mbox{flies}} & \mc{1}{\mbox{like}} & \mc{1}{\mbox{bananas}}\\
\uline{1} & \uline{1} & \uline{1} & \uline{1}\\
\mc{1}{NP} & \mc{1}{NP \bs NP} & \mc{1}{(S\bs NP)/NP} &\mc{1}{NP}\\
\bapply{2} & \fapply{2}\\
 \mc{2}{NP}  & \mc{2}{S\bs NP}\\
\bapply{4}\\
\mc{4}{S}\\
}
}};

{
\fboxrule=2pt%border thickness
\unexplorednode{fruit-flies-like-bananas2}{-4}{-5}{
\deriv{4}{
\mc{1}{\mbox{Fruit}} &\mc{1}{\mbox{flies}} & \mc{1}{\mbox{like}} & \mc{1}{\mbox{bananas}}\\
\uline{1} & \uline{1} & \uline{1} & \uline{1}\\
\mc{1}{NP} & \mc{1}{S \bs NP} & \mc{1}{(S\bs S)/NP} &\mc{1}{NP}\\
\bapply{2} & \fapply{2}\\
 \mc{2}{S}  & \mc{2}{S\bs S}\\
\bapply{4}\\
\mc{4}{S}\\
}
}};
\forestedge{start}{150}{-90}{fruit};
\forestedge{start}{90}{-90}{flies};
\forestedge{start}{30}{-90}{like};
\forestedge{start}{15}{-90}{bananas};
\forestedge{like}{90}{-90}{like-bananas};
\forestedge{bananas}{90}{-90}{like-bananas};
\forestedge{fruit}{90}{-90}{fruit-flies};
\forestedge{flies}{90}{-90}{fruit-flies};
\forestedge{fruit-flies}{60}{-120}{fruit-flies-like-bananas};
\forestedge{like-bananas}{90}{-120}{fruit-flies-like-bananas};

\frontieredge{start}{-165}{60}{fruit2};
\frontieredge{start}{-120}{60}{flies2};
\frontieredge{start}{-15}{90}{like2};
\frontieredge{start}{160}{-90}{flies3};
\unexplorededge{fruit2}{120}{-120}{fruit-flies3};
\unexplorededge{flies3}{90}{-120}{fruit-flies3};
\unexplorededge{fruit-flies3}{90}{-60}{fruit-flies-like-bananas3};
\unexplorededge{like-bananas}{150}{-60}{fruit-flies-like-bananas3};

\unexplorededge{like2}{-90}{75}{like-bananas2};
\unexplorededge{bananas}{0}{75}{like-bananas2};

\unexplorededge{fruit2}{-90}{150}{fruit-flies2};
\unexplorededge{flies2}{-90}{150}{fruit-flies2};

\unexplorededge{fruit-flies2}{-90}{-10}{fruit-flies-like-bananas2};
\unexplorededge{like-bananas2}{-90}{-10}{fruit-flies-like-bananas2};
\end{tikzpicture}
}
}
\caption{\footnotesize The search space in this work, with one node for each \\ partial parse.}
\label{fig:forest-hypergraph}
\end{subfigure}
\caption{\label{fig:search} Illustrations of CCG parsing as hypergraph search, showing partial views of the search space. Weighted hyperedges from child nodes to a parent node represent rule productions scored by a parsing model. A path starting at $\emptyset$, for example the set of bolded hyperedges, represents the derivation of a parse. During decoding, we find the highest scoring path to a complete parse. Both figures show an ideal exploration that efficiently finds the optimal path. Figure~\ref{fig:chart-hypergraph} depicts the traditional search space, and Figure~\ref{fig:forest-hypergraph} depicts the search space in this work. Hyperedge scores can only depend on neighboring nodes, so our model can condition on the entire parse structure, at the price of an exponentially larger search space.
}
\end{figure*}
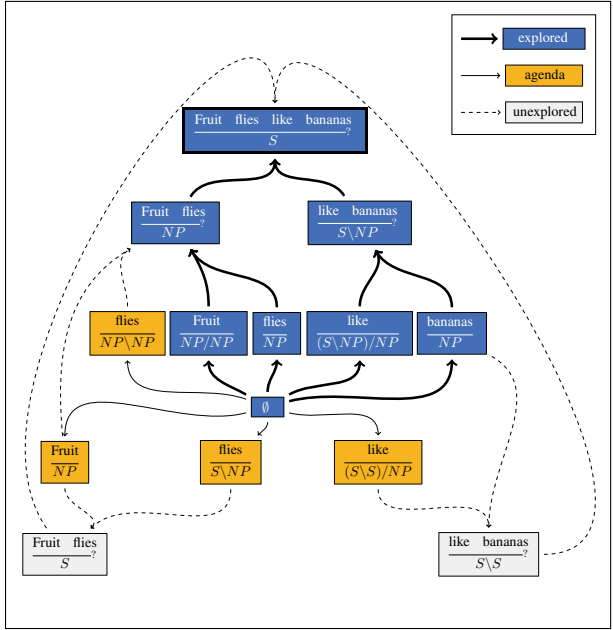
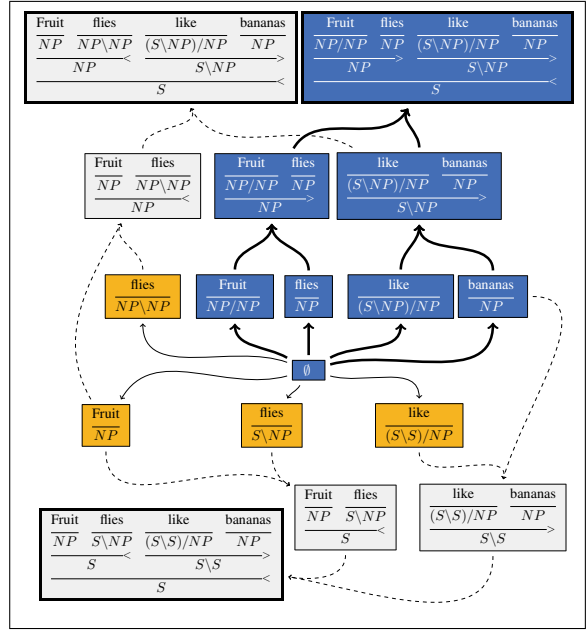

%and the space of structures is too large to be explored exhaustively---meaning that existing work has been forced to use greedy inference techniques such as beam search~\cite{vinyals:2014,dyer:2015} and re-ranking~\cite{socher:2013}.Therefore, solutions returned by decoding may not be the highest scoring according to the model.

To enable learning of global representations, we modify the parser to search directly in the space of all possible parse trees with no dynamic programming. Optimality guarantees come from \astar search, which provides a certificate of optimality if run to completion with a heuristic that is a bound on the future cost.
Generalizing \astar to global models is challenging; these models also break the locality assumptions used to efficiently compute existing \astar heuristics~\cite{klein:2003,lewis:2014b}.
%, which were designed for CKY-style chart parsers. 
Rather than directly replacing local models, we show that they can simply be augmented by adding a score from a global model that is constrained to be non-positive and has a trivial upper bound of zero. The global model, in effect, only needs to model the remaining non-local phenomena. %, which can be a much easier problem in practice.
In our experiments, we use a recent factored \astar CCG parser ~\cite{lewis:2016} for the local model, and we train a Tree-LSTM~\cite{tai:2015} to model global structure.

%it is not possible to collapse parses into equivalence classes, so any optimal decoding algorithm must explore a super-exponential number of parses in the worst case.
%To decode efficiently, we use an \astar algorithm, which can efficiently explore unbounded spaces if it is given an informative upper bound of future cost.
%In this super-exponential search space, we can hope to efficiently find the optimal parse with an \astar parser if (1) we have access to an informative \astar heuristic and (2) the model is confident, allowing the search to progress quickly towards the highest scoring full parse. 
%\astar parsing has previously been used with factored models which allow informative heuristics \cite{klein:2003}---however, these techniques cannot be applied to global models because the heuristic computations require the same local decompositions used for CKY-style parsing. 

%Our solution is to divide the modeling responsibilities between a local component and a monotonically decreasing global component. The local component has limited expressivity, but informative upper bounds for it can be easily computed, thus providing an admissible \astar heuristic to guide the search. The global component, for which we cannot compute an informative upper bound, widens the gap between the \astar heuristic and the true outside score, but only needs to provide corrections for phenomena that the local component cannot model.

Finding a model that achieves these \astar guarantees in practice is a challenging learning problem. 
Traditional structured prediction objectives focus on ensuring that the gold parse has the highest score~\cite{collins:2002,huang:2012}. This condition is insufficient in our case, since it does not guarantee that the search will terminate in sub-exponential time. We instead introduce a new objective that optimizes efficiency as well as accuracy. 
Our loss function is defined over states of the \astar search agenda, and it penalizes the model whenever the top agenda item is not a part of the gold parse. Minimizing this loss encourages the model to return the correct parse as quickly as possible.
The combination of global representations and optimal decoding enables our parser to achieve state-of-the-art accuracy for Combinatory Categorial Grammar (CCG) parsing. Despite being intractable in the worst case, the parser in practice is highly efficient. 
%obtains certificates of optimality for 99.8\% of held-out sentences---and outperforms beam search in both speed and accuracy. 
It finds optimal parses for 99.9\% of held out sentences while exploring just 190 subtrees on average---allowing it to outperform beam search in both speed and accuracy.

\section{Overview}
\label{sec:overview}
\paragraph{Parsing as hypergraph search}
Many parsing algorithms can be viewed as a search problem, where parses are specified by paths through a hypergraph. 

A node $y$ in this hypergraph is a labeled span, representing structures within a parse tree, as shown in Figure~\ref{fig:search}. Each hyperedge $e$ in the hypergraph represents a rule production in a parse. The head node of the hyperedge \Funct{head}{e} is the parent of the rule production, and the tail nodes of the hyperedge are the children of the rule production. For example, consider the hyperedge in Figure~\ref{fig:forest-hypergraph} whose head is \textit{like bananas}. This hyperedge represents a forward application rule applied to its tails, \textit{like} and \textit{bananas}.

To define a path in the hypergraph, we first include a special start node $\emptyset$ that represents an empty parse. $\emptyset$ has outgoing hyperedges that reach every leaf node, representing assignments of labels to words (supertag assignments in Figure~\ref{fig:search}). We then define a path to be a set of hyperedges $E$ starting at $\emptyset$ and ending at a single destination node. A path therefore specifies the derivation of the parse constructed from the labeled spans at each node. For example, in Figure~\ref{fig:search}, the set of bolded hyperedges form a path deriving a complete parse. 

Each hyperedge $e$ is weighted by a score $s(e)$ from a parsing model. The score of a path $E$ is the sum of its hyperedge scores:
{
\setlength{\abovedisplayskip}{6pt}
\setlength{\belowdisplayskip}{6pt}
\begin{align*}
g(E) &= \sum_{e \in E} s(e)
\end{align*}
}
Viterbi decoding is equivalent to finding the highest scoring path that forms a complete parse.

\paragraph{Search on parse forests}
Traditionally, the hypergraph represents a packed parse chart. In this work, our hypergraph instead represents a \emph{forest} of parses. Figure~\ref{fig:search} contrasts the two representations. 

In the parse chart, labels on the nodes represent local properties of a parse, such as the category of a span in Figure~\ref{fig:chart-hypergraph}. As a result, multiple parses that contain the same property include the same node in their path, (e.g. the node spanning the phrase \textit{Fruit flies} with category \texttt{NP}). The number of nodes in this hypergraph is polynomial in the sentence length, permitting exhaustive exploration (e.g. CKY parsing). However, the model scores can only depend on local properties of a parse. We refer to these models as \emph{locally factored} models.

In contrast, nodes in the parse forest are labeled with entire subtrees, as shown in Figure~\ref{fig:forest-hypergraph}. For example, there are two nodes spanning the phrase \textit{Fruit flies} with the same category \texttt{NP} but different internal substructures. While the parse forest requires an exponential number of nodes in the hypergraph, the model scores can depend on entire subtrees.

\paragraph{\astar parsing}
\astar parsing has been successfully applied in locally factored models~\cite{klein:2003,lewis:2014b,lewis:2015,lewis:2016}. We present a special case of \astar parsing that is conceptually simpler, since the parse forest constrains each node to be reachable via a unique path. During exploration, we maintain the unique (and therefore highest scoring) path to a hyperedge $e$, which we define as $\Funct{path}{e}$.

Similar to the standard \astar search algorithm, we maintain an agenda $\agenda$ of hyperedges to explore and a forest $\forest$ of explored nodes that initially contains only the start node $\emptyset$. 

Each hyperedge $e$ in the agenda is sorted by the sum of its inside score $g(\Funct{path}{e})$ and an admissible heuristic $h(e)$. A heuristic $h(e)$ is admissible if it is an upper bound of the sum of hyperedge scores leading to any complete parse reachable from $e$ (the Viterbi outside score). %It is important to choose an informative heuristic, since tighter heuristics lead to more efficient search.
The efficiency of the search improves when this bound is tighter.

At every step, the parser removes the top of the agenda, $e_{max} =  \argmax_{e \in \agenda}(g(\Funct{path}{e}) + h(e))$. $e_{max}$ is expanded by combining $\Funct{head}{e_{max}}$ with previously explored parses from $\forest$ to form new hyperedges. These new hyperedges are inserted into $\agenda$, and $\Funct{head}{e_{max}}$ is added it to $\forest$. We repeat these steps until the first complete parse $y^*$ is explored. The bounds provided by $h(e)$ guarantee that the path to $y^*$ has the highest possible score. Figure~\ref{fig:forest-hypergraph} shows an example of the agenda and the explored forest at the end of perfectly efficient search, where only the optimal path is explored. 

\paragraph{Approach}
The enormous search space described above presents a challenge for an \astar parser, since computing a tight and admissible heuristic is difficult when the model does not decompose locally. 

Our key insight in addressing this challenge is that existing locally factored models with an informative \astar heuristic can be augmented with a global score (Section~\ref{sec:model}). By constraining the global score to be non-positive, the \astar heuristic from the locally factored model is still admissible.

While the heuristic from the local model offers some estimate of the future cost, the efficiency of the parser requires learning a well-calibrated global score, since the heuristic becomes looser as the global score provides stronger penalties (Section~\ref{sec:learning}).

As we explore the search graph, we incrementally construct a neural network, which computes representations of the parses and allows backpropagation of errors from bad search steps (Section~\ref{sec:inference}).

In the following sections, we present our approach in detail, assuming an existing locally factored model $s_{local}(e)$ for which we can efficiently compute an admissible \astar heuristic $h(e)$.

In the experiments, we apply our model to CCG parsing, using the locally factored model and \astar heuristic from \newcite{lewis:2016}.
\section{Model}
\label{sec:model}

\begin{figure}[t!]
\center

\newcommand\layer[2]{
\draw (#1,#2) rectangle (#1+2,#2+1);
\fill[black!0!white] (#1,#2) rectangle (#1+2,#2+1);
\draw (#1 + 0.5, #2 + .5) circle (0.4);

\pgfmathparse{0.9*rnd+0.3}
\definecolor{MyColor}{rgb}{\pgfmathresult,\pgfmathresult,\pgfmathresult}
\fill[MyColor] (#1 + 0.5, #2 + .5) circle (0.4);

\pgfmathparse{0.9*rnd+0.3}
\definecolor{MyColor}{rgb}{\pgfmathresult,\pgfmathresult,\pgfmathresult}
\draw (#1 + 1.5, #2 + .5) circle (0.4);
\fill[MyColor] (#1 + 1.5, #2 + .5) circle (0.4);
}

\newcommand\mybezier[4]{
    \draw [green]          (#1)--(#2)-- (#3)--(#4);
    \draw[very thick,blue] (#1).. controls (#2) and (#3) .. (#4);
    \draw [fill=red,draw=black]    (#1)node[left] {$P_0$} circle (2pt);
    \draw [fill=red,draw=black]    (#2)node[left] {$P_a$} circle (2pt);
    \draw [fill=red,draw=black]    (#3)node[right]{$P_b$} circle (2pt);
    \draw [fill=red,draw=black]    (#4)node[right]{$P_1$} circle (2pt);
}

\newcommand\terminal[3] {
\node[draw=none] at (#1+3,-1.2) {\Huge #2}; % word
\draw (#1+3,-0.5) -- (#1+3,0); % word to embedding
\layer{#1+2}{0} % word embedding
\draw (#1+3,1) -- (#1+1,2); % word embedding to forward
\draw (#1+3,1) -- (#1+5,2); % word embedding to backward

\layer{#1+0}{2} % forward LSTM
\draw (#1+1,3) -- (#1+3,6); % forward LSTM to hidden
\layer{#1+4}{2} % backward LSTM
\draw (#1+5,3) -- (#1+3,6); % backward LSTM to hidden

\node[draw=none] at (#1-0.3,5.0) {\huge \cf{#3}};% category
\draw (#1-0.3,5.5) -- (#1-0.3,6); % category to embedding
\layer{#1-1.5}{6} % category embedding
\draw (#1+0.5,6.5) -- (#1+2,6.5); % category embedding to hidden

\draw (#1+2,2.5) to [bend left] (#1+8,2.5); % forward right
\draw (#1+2-8,2.5) to [bend left] (#1+8-8,2.5); % forward left
\draw (#1+4,2.5) to [bend right] (#1-2,2.5); % backward left
\draw (#1+4+8,2.5) to [bend right] (#1-2+8,2.5); % backward right

\layer{#1+2}{6} % hidden
}

\newcommand\nonterminal[7] {
\layer{#1}{#2}
\draw (#3,#4) -- (#1+1,#2); % left hidden to hidden
\draw (#5,#6) -- (#1+1,#2); % right hidden to hidden

\node[draw=none] at (#1-2.5,#2-1.0) {\huge \cf{#7}};% category
\layer{#1-3.5}{#2} % category embedding
\draw (#1-2.5,#2-0.5) -- (#1-2.5,#2); % category to embedding
\draw (#1-1.5,#2+0.5) -- (#1,#2+0.5); % category embedding to hidden
}

\frame{
\scalebox{0.23} {
\begin{tikzpicture}[->,line width=1.5mm]

\clip(-10,-2) rectangle (24,15);

\terminal{-8}{Fruit}{NP/NP}
\terminal{0}{flies}{NP}
\terminal{8}{like}{(S \bs NP)/NP}
\terminal{16}{bananas}{NP}

\nonterminal{5}{13}{-1}{10}{15}{10}{S}
\nonterminal{-2}{9}{-5}{7}{3}{7}{NP}
\nonterminal{14}{9}{11}{7}{19}{7}{S \bs NP}
\end{tikzpicture}
\hspace{-0.6cm}
}
}
\setlength{\belowcaptionskip}{-5pt}
\caption{\label{fig:lstm} Visualization of the Tree-LSTM which computes vector embeddings for each parse node. The leaves of the Tree-LSTM are connected to a bidirectional LSTM over words, encoding lexical information within and outside of the parse.}
\end{figure}
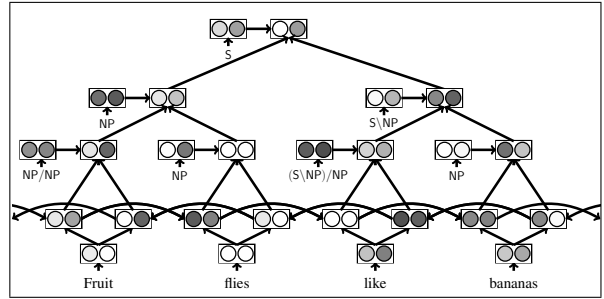

Our model scores a hyperedge $e$ by combining the score from the local model with a global score that conditions on the entire parse at the head node:
{
\setlength{\abovedisplayskip}{6pt}
\setlength{\belowdisplayskip}{6pt}
\begin{align*}
s(e) &= s_{local}(e) + s_{global}(e)
\end{align*}
}\noindent
In $s_{global}(e)$, we first compute a hidden representation encoding the parse structure of $y = \Funct{head}{e}$. We use a variant of the Tree-LSTM~\cite{tai:2015} connected to a bidirectional LSTM~\cite{hochreiter:1997} at the leaves.
The combination of linear and tree LSTMs allows the hidden representation of partial parses to condition on both the partial structure and the full sentence.
Figure~\ref{fig:lstm} depicts the neural network that computes the hidden representation for a parse.

Formally, given a sentence $\langle w_1, w_2, \ldots, w_n \rangle$, we compute hidden states $h_t$ and cell states $c_t$ in the forward LSTM for $1 < t \le n$:
{
\setlength{\abovedisplayskip}{6pt}
\setlength{\belowdisplayskip}{6pt}
\begin{align*}
i_t=&\sigma(W_i [c_{t-1},h_{t-1},x_t] + b_i)\\
o_t=&\sigma(W_o [\tilde{c}_{t},h_{t-1},x_t] + b_o)\\
\tilde{c_{t}}=&\tanh(W_c[h_{t-1},x_t]+b_c)\\
c_t=&i_t\circ \tilde{c}_{t}+(\mathbf{1}-i_t) \circ c_{t-1}\\
h_t=&o_t\circ \tanh(c_t)
\end{align*}
}\noindent
where $\sigma$ is the logistic
sigmoid, $\circ$ is the component-wise product, and $x_t$ denotes a learned word embedding for $w_t$. We also construct a backward LSTM, which produces the analogous hidden and cell states starting at the end of the sentence, which we denote as $c'_t$ and $h'_t$ respectively. The start and end latent states, $c_{-1}$, $h_{-1}$, $c'_{n+1}$, and $h'_{n+1}$, are learned embeddings. This variant of the LSTM includes peephole connections and couples the input and forget gates.

The bidirectional LSTM over the words serves as a base case when we recursively compute a hidden representation for the parse $y$ using the tree-structured generalization of the LSTM:
{
\setlength{\abovedisplayskip}{6pt}
\setlength{\belowdisplayskip}{6pt}
\begin{align*}
i_y &= \sigma(W^R_i [c_l, h_l, c_r, h_r, x_y] + b^R_i)\\
f_{y} &= \sigma(W^R_{f} [c_l, h_l, c_r, h_r, x_y] + b^R_{f})\\
o_y &= \sigma(W^R_o[\widetilde{c}_y, h_l, h_r, x_y] + b^R_o)\\
c_{lr} &= f_y\circ c_{l}+(\mathbf{1}-f_y)  \circ c_{r}\\
%f'_{y} &= \mathbf{1} - f_{y}\\
\widetilde{c}_y &= \tanh(W^R_c [h_l, h_r, x_y] + b^R_c)\\
%c_y &= i_y \circ \widetilde{c}_y +  (\mathbf{1} - i_y) \circ (f_{y} \circ c_l + f'_{y} \circ c_r) \\
c_y &= i_y \circ \widetilde{c}_y +  (\mathbf{1} - i_y) \circ c_{lr} \\
h_y &= o_y \circ \tanh(c_y)
\end{align*}
}\noindent
where the weights and biases are parametrized by the rule $R$ that produces $y$ from its children, and $x_y$ denotes a learned embedding for the category at the root of $y$. For example, in CCG, the rule would correspond to the CCG combinator, and the label would correspond to the CCG category.

We assume that nodes are binary, unary, or leaves. Their left and right latent states, $c_l$, $h_l$, $c_r$, and $h_r$ are defined as follows:
\begin{itemize}[noitemsep]
\item In a binary node, $c_l$ and $h_l$ are the cell and hidden states of the left child, and $c_r$ and $h_r$ are the cell and hidden states of the right child.
\item In a unary node, $c_l$ and $h_l$ are learned embeddings, and $c_r$ and $h_r$ are the cell and hidden states of the singleton child.
\item In a leaf node, let $w$ denote the index of the corresponding word. Then $c_l$ and $h_l$ are $c_{w}$ and $h_{w}$ from the forward LSTM, and $c_r$ and $h_r$ are $c'_{w}$ and $h'_{w}$ from the backward LSTM.
\end{itemize}

The cell state of the recursive unit is a linear combination of the intermediate cell state $\widetilde{c}_y$, the left cell state $c_l$, and the right cell state $c_r$. To preserve the normalizing property of coupled gates, we perform coupling in a hierarchical manner: the input gate $i_y$ decides the weights for $\widetilde{c}_y$, and the forget gate $f_y$ shares the remaining weights between $c_l$ and $c_r$.

Given the hidden representation $h_y$ at the root, we score the global component as follows:
{
\setlength{\abovedisplayskip}{6pt}
\setlength{\belowdisplayskip}{6pt}
\begin{align*}
s_{global}(e) = \log(\sigma(W \cdot h_y))
\end{align*}
}
This definition of the global score ensures that it is non-positive---an important property for inference.

\section{Inference}
\label{sec:inference}
Using the hyperedge scoring model $s(e)$ described in Section~\ref{sec:model}, we can find the highest scoring path that derives a complete parse tree by using the \astar parsing algorithm described in Section~\ref{sec:overview}.

\paragraph{Admissible \astar heuristic}
Since our full model adds non-positive global scores to the existing local scores, path scores under the full model cannot be greater than path scores under the local model. Upper bounds for path scores under the local model also hold for path scores under the full model, and we simply reuse the \astar heuristic from the local model to guide the full model during parsing without sacrificing optimality guarantees.

\paragraph{Incremental neural network construction}
The recursive hidden representations used in $s_{global}(e)$ can be computed in constant time during parsing. When scoring a new hyperedge, its children must have been previously scored. Instead of computing the full recursion, we reuse the existing latent states of the children and compute $s_{global}(e)$ with an incremental forward pass over a single recursive unit in the neural network. By maintain the latent states of each parse, we incrementally build a single DAG-structured LSTM mirroring the explored subset of the hypergraph. This not only enables quick forward passes during decoding, but also allows backpropagation through the search space after decoding, which is crucial for efficient learning (see Section~\ref{sec:learning}).

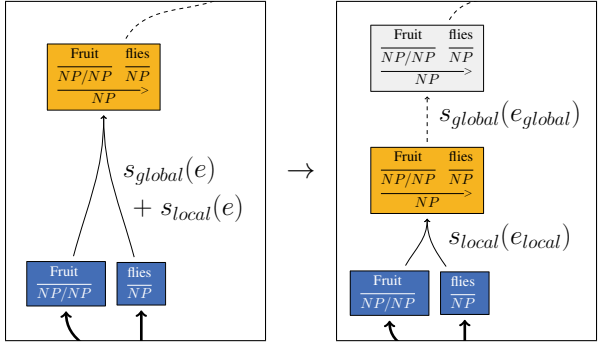
\begin{figure}[t!]
\begin{subfigure}{0.45\columnwidth}
\frame{
\scalebox{0.5}{
\begin{tikzpicture}
\clip(-3.5,1) rectangle (3, 10);
\forestnode{fruit}{-2}{2.5}{
\deriv{2}{
\mc{1}{\mbox{Fruit}}\\
\uline{1}\\
\mc{1}{NP/NP}\\
}
};
\forestnode{flies}{0}{2.5}{
\deriv{2}{
\mc{1}{\mbox{flies}}\\
\uline{1}\\
\mc{1}{NP}\\
}
};
\frontiernode{fruit-flies}{-1}{8}{
\deriv{2}{
\mc{1}{\mbox{Fruit}} & \mc{1}{\mbox{flies}}\\
\uline{1} & \uline{1}\\
\mc{1}{NP/NP} & \mc{1}{NP}\\
 \fapply{2}\\
 \mc{2}{NP}\\
}
};
\unexplorednode{fruit-flies-like-bananas}{3.5}{12}{
\deriv{4}{
\mc{1}{\mbox{Fruit}} &\mc{1}{\mbox{flies}} & \mc{1}{\mbox{like}} & \mc{1}{\mbox{bananas}}\\
\uline{1} & \uline{1} & \uline{1} & \uline{1}\\
\mc{1}{NP/NP} & \mc{1}{NP} & \mc{1}{(S\bs NP)/NP} &\mc{1}{NP}\\
\fapply{2} & \fapply{2}\\
 \mc{2}{NP}  & \mc{2}{S\bs NP}\\
\bapply{4}\\
\mc{4}{S}\\
}
};
\forestedge{start}{150}{-90}{fruit};
\forestedge{start}{90}{-90}{flies};
\frontieredge{fruit}{75}{-90}{fruit-flies};
\frontieredge{flies}{105}{-90}{fruit-flies};
\unexplorededge{fruit-flies}{60}{-120}{fruit-flies-like-bananas};
\node at (0.75, 5.5) (a) {\huge $s_{global}(e)$};
\node at (1.25, 4.5) (a) {\huge $+\;s_{local}(e)$};
\end{tikzpicture}
}
}
\end{subfigure}
\hspace{0.05cm}\textbf{$\rightarrow$}\hspace{0.21cm}
\begin{subfigure}{0.45\columnwidth}
\frame{
\scalebox{0.5}{
\begin{tikzpicture}
\clip(-3.3,1) rectangle (3.2, 10);
\forestnode{fruit}{-2}{2.25}{
\deriv{2}{
\mc{1}{\mbox{Fruit}}\\
\uline{1}\\
\mc{1}{NP/NP}\\
}
};
\forestnode{flies}{0}{2.25}{
\deriv{2}{
\mc{1}{\mbox{flies}}\\
\uline{1}\\
\mc{1}{NP}\\
}
};
\frontiernode{fruit-flies}{-1}{5.25}{
\deriv{2}{
\mc{1}{\mbox{Fruit}} & \mc{1}{\mbox{flies}}\\
\uline{1} & \uline{1}\\
\mc{1}{NP/NP} & \mc{1}{NP}\\
 \fapply{2}\\
 \mc{2}{NP}\\
}
};
\unexplorednode{fruit-flies2}{-1}{8.5}{
\deriv{2}{
\mc{1}{\mbox{Fruit}} & \mc{1}{\mbox{flies}}\\
\uline{1} & \uline{1}\\
\mc{1}{NP/NP} & \mc{1}{NP}\\
 \fapply{2}\\
 \mc{2}{NP}\\
}
};
\unexplorednode{fruit-flies-like-bananas}{3.5}{12}{
\deriv{4}{
\mc{1}{\mbox{Fruit}} &\mc{1}{\mbox{flies}} & \mc{1}{\mbox{like}} & \mc{1}{\mbox{bananas}}\\
\uline{1} & \uline{1} & \uline{1} & \uline{1}\\
\mc{1}{NP/NP} & \mc{1}{NP} & \mc{1}{(S\bs NP)/NP} &\mc{1}{NP}\\
\fapply{2} & \fapply{2}\\
 \mc{2}{NP}  & \mc{2}{S\bs NP}\\
\bapply{4}\\
\mc{4}{S}\\
}
};
\forestedge{start}{150}{-90}{fruit};
\forestedge{start}{90}{-90}{flies};
\frontieredge{fruit}{60}{-90}{fruit-flies};
\frontieredge{flies}{125}{-90}{fruit-flies};
\unexplorededge{fruit-flies}{90}{-90}{fruit-flies2};
\unexplorededge{fruit-flies2}{60}{-120}{fruit-flies-like-bananas};
\node at (1.2, 7) (a) {\huge $s_{global}(e_{global})$};
\node at (1.2, 3.75) (a) {\huge $s_{local}(e_{local})$};
\end{tikzpicture}
}
}
\end{subfigure}
\caption{\label{fig:lazy} The hyperedge on the left requires computing both the local and global score when placed on the agenda. Splitting the hyperedge, as shown on the right, saves expensive computation of the global score if the local score alone indicates that the parse is not worth exploring.}
\vspace{-5pt}
\end{figure}

\paragraph{Lazy global scoring}
The global score is expensive to compute. We introduce an optimization to avoid computing it when provably unnecessary.
We split each hyperedge $e$ into two successive hyperedges, $e_{local}$ and $e_{global}$, as shown in Figure~\ref{fig:lazy}. The score for $e$, previously $s(e) = s_{local}(e) + s_{global}(e)$, is also split between the two new hyperedges:
{
\setlength{\abovedisplayskip}{6pt}
\setlength{\belowdisplayskip}{6pt}
\begin{align*}
s(e_{local}) &= s_{local}(e_{local})\\
s(e_{global}) &= s_{global}(e_{global})
\end{align*}
}
Intuitively, this transformation requires \astar to verify that the local score is good enough before computing the global score, which requires an incremental forward pass over a recursive unit in the neural network. In the example, this involves first summing the supertag scores of \textit{Fruit} and \textit{flies} and inserting the result back into the agenda. The score for applying the forward application rule to the recursive representations is only computed if that item appears again at the head of the agenda. In practice, the lazy global scoring reduces the number of recursive units by over 91\%, providing a 2.4X speed up.

\section{Learning}
\label{sec:learning}
During training (Algorithm~\ref{alg:learn}), we assume access to sentences labeled with gold parse trees $\hat{y}$ and gold derivations $\hat{E}$. The gold derivation $\hat{E}$ is a path from $\emptyset$ to $\hat{y}$ in the parse forest.

%Traditional parsing approaches have a tractable upper bound on the time taken for inference, and are trained to maximize accuracy.
%Our model has no such guarantees, so needs to be trained to search both efficiently and accurately.
%The space of possible parses is far larger than can be searched exhaustively, so the model must be trained to efficiently search for $\hat{y}$ given the sentence. 
\astar search with our global model is not guaranteed to terminate in sub-exponential time.
This creates challenges for learning---for example, it is not possible in practice to use the standard structured perceptron update~\cite{collins:2002}, because the search procedure rarely terminates early in training.
Other common loss functions assume inexact search \cite{huang:2012}, and do not optimize efficiency.

Instead, we optimize a new objective that is tightly coupled with the search procedure. During parsing, we would like hyperedges from the gold derivation to appear at the top of the agenda $\agenda$. When this condition does not hold, \astar is searching inefficiently, and we refer to this as a \emph{violation} of the agenda, which we formally define as:
{
\setlength{\abovedisplayskip}{6pt}
\setlength{\belowdisplayskip}{6pt}
\begin{align*}
v(\hat{E}, \agenda) &= \max_{e \in \agenda}(g(\Funct{path}{e})+ h(e)) \\
&\;\;\;\;- \max_{e \in \agenda \cap \hat{E}}(g(\Funct{path}{e}) + h(e))
\end{align*}
}\noindent
where $g(\Funct{path}{e})$ is the score of the unique path to $e$, and $h(e)$ is the \astar heuristic. If all violations are zero, we find the gold parse without exploring any incorrect partial parses---maximizing both accuracy and efficiency. Figure~\ref{fig:forest-hypergraph} shows such a case---if any other nodes were explored, they would be violations. 

In existing work on violation-based updates, comparisons are only made between derivations with the same number of steps~\cite{huang:2012,clark:2015}---whereas our definition allows subtrees of arbitrary spans to compete with each other, because hyperedges are not explored in a fixed order. Our violations also differ from Huang et al.'s in that we optimize efficiency as well as accuracy.

We define loss functions over these violations, which are minimized to encourage correct and efficient search. During training, we parse each sentence until either the gold parse is found or we reach computation limits. We record $\violation$, the list of non-zero violations of the agenda $\agenda$ observed:
{
\setlength{\abovedisplayskip}{6pt}
\setlength{\belowdisplayskip}{6pt}
\begin{align*}
\violation &= \langle v(\hat{E}, \agenda) \mid v(\hat{E}, \agenda) > 0 \rangle
\end{align*}
}
\begin{table}[t!]
\begin{center}
\scalebox{0.85}{
\setlength{\tabcolsep}{5pt}
\def\arraystretch{1}
\begin{tabular}{|l|l|}
\hline
\textbf{Update} & \textbf{\Funct{Loss}{$\violation$}} \\
\hline
%Greedy & $\violation_{\argmin_t \violation_t > 0}$ \\
Greedy & $\violation_1$ \\
%Early & $\violation_T$ \\
Max violation & $\max_{t=1}^{T} \violation_t$ \\
All violations & $\sum_{t=1}^{T} \violation_t$ \\
\hline
\end{tabular}
}
\caption{\label{tab:update-definitions} Loss functions optimized by the different update methods. The updates depend on the list of $T$ non-zero violations, $\violation = \langle \violation_1, \violation_2, \ldots, \violation_T \rangle$, as defined in Section~\ref{sec:learning}.}
\end{center}
\vspace{-5pt}
\end{table}
We can optimize several loss functions over $\violation$, as defined in Table~\ref{tab:update-definitions}. The greedy and max-violation updates are roughly analogous to the violation-fixing updates proposed by \newcite{huang:2012}, but adapted to exact agenda-based parsing.
%The gold parse is unreachable only when we terminate the search, since we do not prune the search space. Therefore, the loss of the early update corresponds to the last violation.
We also introduce a new \emph{all-violations} update, which minimizes the sum of all observed violations. The all-violations update encourages correct parses to be explored early (similar to the greedy update) while being robust to parses with multiple deviations from the gold parse (similar to the max-violation update).

The violation losses are optimized with subgradient descent and backpropagation. For our experiments, $s_{local}(e)$ and $h(e)$ are kept constant. Only the parameters $\theta$ of $s_{global}(e)$ are updated. Therefore, a subgradient of a violation $v(\hat{E}, \agenda)$ can be computed by summing subgradients of the global scores.
{
\setlength{\abovedisplayskip}{6pt}
\setlength{\belowdisplayskip}{6pt}
\begin{align*}
\frac{\partial v(\hat{E}, \agenda)}{\partial \theta} &= \hspace{-0.5cm} \sum_{e \in \Funct{path}{e_{max}}} \hspace{-0.5cm} \frac{\partial s_{global}(e)}{\partial \theta} - \hspace{-0.5cm}\sum_{e \in \Funct{path}{\hat{e}_{max}}}\hspace{-0.5cm} \frac{\partial s_{global}(e)}{\partial \theta}
\end{align*}
}\noindent
where $e_{max}$ denotes the hyperedge at the top of the agenda $\agenda$ and $\hat{e}_{max}$ denotes the hyperedge in the gold derivation $\hat{E}$ that is closest to the top of $\agenda$.
\begin{algorithm}[t!]
  \footnotesize
  \caption{Violation-based learning algorithm
    \label{alg:learn}}
  \textbf{Definitions} $D$ is the training data containing input sentences $x$ and gold derivations $\hat{E}$. $e$ variables denote scored hyperedges. \Funct{tag}{$x$} returns a set of scored pre-terminals for every word.  \Funct{add}{$\forest$, $y$} adds partial parse $y$ to forest $\forest$. \Funct{rules}{$\forest$, $y$} returns the set of scored hyperedges that can be created by combining $y$ with entries in $F$. \Funct{size\_ok}{$\forest$, $\agenda$} returns whether the sizes of the forest and agenda are within predefined limits.
  \begin{algorithmic}[1]
    \Function{violations}{$\hat{E}, x, \theta$}
      \Let{$\violation$}{$\emptyset$}\Comment{Initialize list of violations $\violation$}
      \Let{$\forest$}{$\emptyset$}\Comment{Initialize forest $\forest$}
      \Let{$\agenda$}{$\emptyset$}\Comment{Initialize agenda $\agenda$}
      \For{$e \in \Funct{tag}{x}$}
        \State \Funct{push}{$\agenda,e$}
      \EndFor
      \While{$|\agenda \cap \hat{E}| > 0~\textbf{and}~ \Funct{size\_ok}{\forest, \agenda}$} 
        \If{$v(\hat{E}, \agenda) > 0$}
          \State \Funct{append}{$\violation,v(\hat{E}, \agenda)$}\Comment{Record violation}
        \EndIf
        \Let{$e_{max}$}{\Funct{extract\_max}{$\agenda$}}\Comment{Pop agenda}
        \State \Funct{add}{$\forest,\Funct{head}{e_{max}}$}\Comment{Explore hyperedge}
        \For {$e \in \Funct{rules}{\forest,\Funct{head}{e_{max}}, \theta}$}
            \State \Funct{push}{$\agenda,e$}\Comment{Expand hyperedge}
        \EndFor
      \EndWhile
      \State \Return{$\violation$}
    \EndFunction
    \\
    \Function{learn}{$D$}
      \For{$i=1$ to $T$}
        \For{$x,\hat{E} \in D$}
          \Let{$\violation$}{$\Funct{violations}{\hat{E}, x, \theta$}}
          \Let{$L$}{$\Funct{loss}{\violation}$}
          \Let{$\theta$}{$\Funct{optimize}{L, \theta}$}
        \EndFor
      \EndFor
      \State \Return{$\theta$}
    \EndFunction
  \end{algorithmic}
\end{algorithm}
\vspace{-30pt}
\section{Experiments}
\subsection{Data}
We trained our parser on Sections 02-21 of CCGbank \cite{hockenmaier:2007}, using Section 00 for development and Section 23 for test. To recover a single gold derivation for each sentence to use during training, we find the right-most branching parse that satisfies the gold dependencies.

\subsection{Experimental Setup}
For the local model, we use the \emph{supertag-factored} model of ~\newcite{lewis:2016}.
Here, $s_{local}(e)$ corresponds to a supertag score if a $\Funct{head}{e}$ is a leaf and zero otherwise.
The outside score heuristic is computed by summing the maximum supertag score for every word outside of each span. In the reported results, we back off to the supertag-factored model after the forest size exceeds 500,000, the agenda size exceeds 2 million, or we build more than 200,000 recursive units in the neural network.

%We use the state-of-the-art bidirectional-LSTM supertagger from~\newcite{lewis:2016} for the locally-factored model. In this setup, $s_{local}(e)$ corresponds to a supertag score if a $\Funct{parent}{e}$ is a leaf and zero otherwise. 

Our full system is trained with all-violations updates. During training, we lower the forest size limit to 2000 to reduce training times. The model is trained for 30 epochs using ADAM~\cite{kingma:2016}, and we use early stopping based on development F1. The LSTM cells and hidden states have 64 dimensions. We initialize word representations with pre-trained 50-dimensional embeddings from~\newcite{turian:2010}. Embeddings for categories have 16 dimensions and are randomly initialized. We also apply dropout with a probability of 0.4 at the word embedding layer during training. Since the structure of the neural network is dynamically determined, we do not use mini-batches. The neural networks are implemented using the CNN library,\footnote{\url{https://github.com/clab/cnn}} and the CCG parser is implemented using the EasySRL library.\footnote{\url{https://github.com/mikelewis0/EasySRL}} The code is available online.\footnote{\url{https://github.com/kentonl/neuralccg}}

\subsection{Baselines}
We compare our parser to several baseline CCG parsers: the C\&C parser~\cite{clark:2007}; C\&C + RNN~\cite{xu:2015}, which is the C\&C parser with an RNN supertagger; \newcite{xu:2016b}, a LSTM shift-reduce  parser; \newcite{vaswani:2016a} who combine a bidirectional LSTM supertagger with a beam search parser using global features \cite{clark:2015}; and \emph{supertag-factored} \cite{lewis:2016}, which uses deterministic \astar decoding and an LSTM supertagging model.
\begin{table}[t!]
\begin{center}
\scalebox{0.9}{
\setlength{\tabcolsep}{5pt}
\begin{tabular}{lllll}
\hline
\textbf{Model} & \textbf{Dev F1} & \textbf{Test F1} \\
\hline
C \& C & 83.8 & 85.2 \\
C \& C + RNN & 86.3 & 87.0 \\
Xu (2016) & 87.5 & 87.8 \\
Vaswani et al. (2016) & 87.8 & 88.3 \\
Supertag-factored & 87.5 & 88.1 \\
\textbf{Global \astar} & \textbf{88.4} & \textbf{88.7}  \\
\hline
\end{tabular}
}
\caption{\label{tab:test} Labeled F1 for CCGbank dependencies on the CCGbank development and test set for our system \textbf{Global \astar} and the baselines.}
\end{center}
\vspace{-5pt}
\end{table}
\subsection{Parsing Results}   
Table~\ref{tab:test} shows parsing results on the test set. 
Our global features let us improve over the supertag-factored model by 0.6 F1.
\newcite{vaswani:2016a} also use global features, but our optimal decoding leads to an improvement of 0.4 F1.

Although we observed an overall improvement in parsing performance, the supertag accuracy was not significantly different after applying the parser.

On the test data, the parser finds the optimal parse for 99.9\% sentences before reaching our computational limits. On average, we parse 27.1 sentences per second,\footnote{We use a single 3.5GHz CPU core.} while exploring only 190.2 subtrees.

\subsection{Model Ablations}
We ablate various parts of the model to determine how they contribute to the accuracy and efficiency of the parser, as shown in Table~\ref{tab:model-ablations}. For each model, the comparisons include the average number of parses explored and the percentage of sentences for which an optimal parse can be found without backing off.

\begin{table}[t!]
\begin{center}
\scalebox{0.85}{
\setlength{\tabcolsep}{5pt}
\begin{tabular}{llll}
\hline
\textbf{Model} & \textbf{Dev F1} & \textbf{Optimal} & \textbf{Explored} \ignore{& \textbf{\# NN}} \\
\hline
Supertag-factored & 87.5 & 100.0\% & 402.5 \ignore{& - }\\
$-$ dynamic program & 87.5 & 97.1\% & 17119.6 \ignore{& - }\\
Span-factored & 87.9 & 99.9\% & 176.5 \ignore{& 494.5} \\
$-$ dynamic program & 87.8 & 99.5\% & 578.5 \ignore{& 1379.7} \\
\textbf{Global \astar} & \textbf{88.4} & 99.8\% & 309.6 \ignore{& 669.5}\\
$-$ lexical inputs & 87.8 & 99.6\% & 538.5 \ignore{& 1321.0} \\
$-$ lexical context & 88.1 & 99.4\% & 610.5 \ignore{& 1502.3}\\
\hline
\end{tabular}
}
\caption{\label{tab:model-ablations} Ablations of our full model (\textbf{Global \astar}) on the development set. \emph{Explored} refers to the size of the parse forest. Results show the importance of global features and lexical information in context.}
\end{center}
\vspace{-5pt}
\end{table}
\paragraph{Structure ablation} We first ablate the global score, $s_{global}(y)$, from our model, thus relying entirely on the local supertag-factors that do not explicitly model the parse structure. This ablation allows dynamic programming and is equivalent to the backoff model (\emph{supertag-factored} in Table~\ref{tab:model-ablations}). Surprisingly, even in the exponentially larger search space, the global model explores \emph{fewer} nodes than the supertag-factored model---showing that the global model efficiently prune large parts of the search space. 
This effect is even larger when not using dynamic programming in the supertag-factored model.
%Without dynamic programming, the supertag-factored model explores almost two magnitudes of parses more than our full model.

\paragraph{Global structure ablation} To examine the importance of global features, we ablate the recursive hidden representation (\emph{span-factored} in Table~\ref{tab:model-ablations}). The model in this ablation decomposes over labels for spans, as in \newcite{durrett:2015}. In this model, the recursive unit uses, instead of latent states from its children, the latent states of the backward LSTM at the start of the span and the latent states of the forward LSTM at the end of the span. Therefore, this model encodes the lexical information available in the full model but does not encode the parse structure beyond the local rule production. While the dynamic program allows this model to find the optimal parse with fewer explorations, the lack of global features significantly hurts its parsing accuracy.

\begin{figure}[t!]
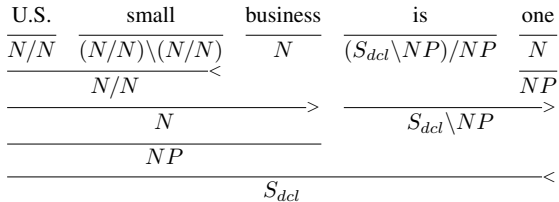

\small
\scalebox{0.85} {
\centering
\deriv{5}{
\mc{1}{\mbox{U.S.}} & \mc{1}{\mbox{small}} & \mc{1}{\mbox{business}} & \mc{1}{\mbox{is}} & \mc{1}{\mbox{one}}\\
\uline{1} & \uline{1} & \uline{1} & \uline{1} & \uline{1}\\
\mc{1}{N/N} & \mc{1}{(N/N)\bs (N/N)} & \mc{1}{N} & \mc{1}{(S_{dcl}\bs NP)/NP} & \mc{1}{N}\\
\bapply{2} &  &  & \uline{1}\\
\mc{2}{N/N} &  &  & \mc{1}{NP}\\
\fapply{3} & \fapply{2}\\
\mc{3}{N} & \mc{2}{S_{dcl}\bs NP}\\
\uline{3}\\
\mc{3}{NP}\\
\bapply{5}\\
\mc{5}{S_{dcl}}\\
}
}
\caption{\label{fig:garden} Example of an incorrect partial parse that appears syntactically plausible in isolation. The full sentence is \textit{`Indeed, for many Japanese trading companies, the favorite \textbf{U.S. small business is one} whose research and development can be milked for future Japanese use.'} The global model heavily penalizes this garden path, thereby avoiding regions that lead to dead ends and allowing the global model to explore fewer nodes. }
\vspace{-5pt}
\end{figure}

\paragraph{Lexical ablation}
We also show lexical ablations instead of structural ablations. We remove the bidirectional LSTM at the leaves, thus delexicalizing the global model. This ablation degrades both accuracy and efficiency, showing that the model uses lexical information to discriminate between parses.

To understand the importance of contextual information, we also perform a partial lexical ablation by using word embeddings at the leaves instead of the bidirectional LSTM, thus propagating only lexical information from within the span of each parse. The degradation in F1 is about half of the degradation from the full lexical ablation, suggesting that a significant portion of the lexical cues comes from the context of a parse. Figure~\ref{fig:garden} illustrates the importance of context with an incorrect partial parse that appears syntactically plausible in isolation. These bottom-up garden paths are typically problematic for parsers, since their incompatibility with the remaining sentence is difficult to recognize until later stages of decoding. However, our global model learns to heavily penalize these garden paths by using the context provided by the bidirectional LSTM and avoid paths that lead to dead ends or bad regions of the search space.

\subsection{Update Comparisons}
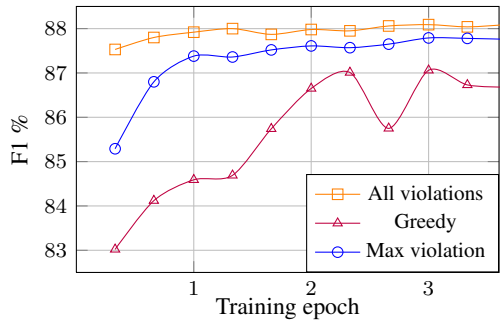
\begin{figure}[t!]
\begin{center}
\begin{small}
\begin{tikzpicture}
\begin{axis}[
    	width=0.9\columnwidth,
	    height=0.64\columnwidth,
	    legend style={at={(1, 0)},anchor=south east,font=\scriptsize},
	    mark options={mark size=2},
		font=\scriptsize,
	    xmin=0.0,xmax=3.6,
   		ymin=82.5, ymax=88.5,
	    xtick={1,2,3},
   	 	ytick={83, 84, 85, 86, 87, 88},
   	 	ymajorgrids=true,
    	xmajorgrids=true,
    	xlabel style={yshift=2.0ex,},
        xlabel=Training epoch,
        ylabel=F1 \%,
    	ylabel style={yshift=-3.0ex,}]
 
    \addplot[smooth,mark=square,orange] plot coordinates {
		(0.33,87.53)
        (0.66,87.80)
        (1.00,87.92)
		(1.33,88.00)
        (1.66,87.87)
        (2.00,87.98)
		(2.33,87.95)
        (2.66,88.06)
        (3.00,88.09)
        (3.33,88.04)
        (3.66,88.09)
    };
    \addlegendentry{All violations}
    
    \addplot[smooth,mark=triangle,purple] plot coordinates {
		(0.33,83.02)
        (0.66,84.12)
        (1.00,84.59)
        (1.33,84.69)
        (1.66,85.74)
        (2.00,86.65)
        (2.33,87.01)
        (2.66,85.75)
        (3.00,87.06)
        (3.33,86.73)
        (3.66,86.68)
    };
    \addlegendentry{Greedy}
    
    %\addplot[smooth,mark=o,blue] plot coordinates {
	%	(0.33,86.99)
    %    (0.66,87.42)
    %    (1.00,87.51)
    %    (1.33,87.62)
    %    (1.66,87.72)
    %    (2.00,87.72)
    %    (2.33,87.75)
    %    (2.66,87.76)
    %    (3.00,87.92)
    %    (3.33,87.75)
    %    (3.66,87.76)
    %};
    %\addlegendentry{Early}
    
    % \addplot[smooth,mark=*,black] plot coordinates {
	%	(0.33,84.52)
    %    (0.66,86.80)
    %    (1.00,86.30)
    %    (1.33,86.30)
    %    (1.66,86.99)
    %    (2.00,86.71)
    %    (2.33,86.74)
    %    (2.66,86.83)
    %    (3.00,86.82)
    %    (3.33,86.88)
    %    (3.66,86.75)
    %};
    %\addlegendentry{Early CRF}
    
    \addplot[smooth,mark=o,blue] plot coordinates {
		(0.33,85.29)
        (0.66,86.80)
        (1.00,87.38)
        (1.33,87.36)
        (1.66,87.52)
        (2.00,87.61)
        (2.33,87.57)
        (2.66,87.65)
        (3.00,87.79)
        (3.33,87.78)
        (3.66,87.76)
    };
    \addlegendentry{Max violation}
    \end{axis}
\end{tikzpicture}
\end{small}
\end{center}
%\vspace{-8pt}
\caption{Learning curves for the first 3 training epochs on the development set when training with different updates strategies. The all-violations update shows the fastest convergence.}
\label{fig:learning}
\vspace{-5pt}

\end{figure}

Table~\ref{tab:update-comparisons} compares the different violation-based learning objectives, as discussed in Section~\ref{sec:learning}. Our novel \emph{all-violation} updates outperform the alternatives. We attribute this improvement to the robustness over poor search spaces, which the greedy update lacks, and the incentive to explore good parses early, which the max-violation update lacks. Learning curves in Figure~\ref{fig:learning} show that the all-violations update also converges more quickly.

\begin{table}[t!]
\begin{center}
\scalebox{0.85}{
\setlength{\tabcolsep}{5pt}
\begin{tabular}{lllll}
\hline
\textbf{Update} & \textbf{Dev F1} & \textbf{Optimal} & \textbf{Explored} \ignore{& \textbf{\# NN}} \\
\hline
Greedy & 87.9 & 99.2\% & 2313.8 \ignore{& 2885.8} \\
%Early & 88.0 & 99.8\% & 258.3 \ignore{& 537.9}\\
Max-violation & 88.1 & 99.9\% & 217.3 \ignore{& 438.1}\\
\textbf{All-violations} & \textbf{88.4} & 99.8\% & 309.6 \ignore{& 669.5} \\
\hline
\end{tabular}
}
\caption{\label{tab:update-comparisons} Parsing results trained with different update methods. Our system uses \textbf{all-violations} updates and is the most accurate.}
\end{center}
%\vspace{-5pt}
\end{table}

% !TEX root =  main.tex

\subsection{Decoder Comparisons}
Lastly, to show that our parser is both more accurate and efficient than other decoding methods, we decode our full model using best-first search, reranking, and beam search. Table~\ref{tab:decoder-comparisons} shows the F1 scores with and without the backoff model, the portion of the sentences that each decoder is able to parse, and the time spent decoding relative to the \astar parser.

In the best-first search comparison, we do not include the informative \astar heuristic, and the parser completes very few parses before reaching computational limits---showing the importance of heuristics in large search spaces. In the reranking comparison, we obtain $n$-best lists from the backoff model and rerank each result with the full model. In the beam search comparison, we use the approach from \newcite{clark:2015} which greedily finds the top-$n$ parses for each span in a bottom-up manner. Results indicate that both approximate methods are less accurate and slower than \astar.

\begin{table}[t!]
\begin{center}
\scalebox{0.85}{
\setlength{\tabcolsep}{5pt}
\begin{tabular}{llll}
\hline
\ignore{\multirow{2}{*}} {\textbf{Decoder}} & \textbf{Dev F1} & \textbf{Dev F1} &  \ignore{\textbf{\# Steps} & \textbf{\# NN} & \textbf{Sentences / s } &} \textbf{Relative}\\
&& \textbf{$-$ backoff} & \textbf{Time}\\
\hline
\textbf{Global \astar} & \textbf{88.4} & 88.4 (99.8\%) & \ignore{309.6 & 669.5 & 12.132 &} 1X\\
%$-$ lazy agenda & 88.4 & 88.1 (99.4\%) & \ignore{251.1 & 2565.5 & 4.9790 &} 2.4X\\
Best-first & 87.5 & 2.8\;\;\;(6.7\%) & \ignore{58464  & 69321.5 & 0.041345 &} 293.4X\\
%10-best reranking & 87.9 & 87.9 (99.7\%) & \ignore{726.2 & 442.3 & 6.62082 &} 1.8X\\
10-best reranking & 87.9 & 87.9 (99.7\%) & \ignore{726.2 & 442.3 & 1.42709 &} 8.5X\\
%100-best reranking & 88.2  &88.0 (99.4\%) & \ignore{3398.9 & 3162.6 & 2.6408 &} 4.6X \\
100-best reranking & 88.2  &88.0 (99.4\%) & \ignore{3398.9 & 3162.6 & 0.16769 &} 72.3X \\
2-best beam search & 88.2 & 85.7 (94.0\%) & \ignore{413.4 & 2139.2 & 6.0895 &} 2.0X \\
4-best beam search & 88.3  & 88.1 (99.2\%) & \ignore{1052.5 & 7185.5 & 1.79982  &} 6.7X \\
8-best beam search &  88.2  & 86.8 (98.1\%) & \ignore{2500.6 & 27690.0 & 0.4619 &} 26.3X\\
\hline
\end{tabular}
}
\caption{\label{tab:decoder-comparisons} Comparison of various decoders using the same model from our full system (\textbf{Global \astar}). We report F1 with and without the backoff model, the percentage of sentences that the decoder can parse, and the time spent decoding relative to \astar.}
\end{center}
\vspace{-5pt}
\end{table}

%\begin{figure*}[ht!]
%\begin{center}
%\setlength{\tabcolsep}{5pt}
%\begin{tabular}{lllll}
%\hline
%\textbf{Decoder} & \textbf{Dev F1 \%} & \textbf{\# Steps} & \textbf{\# NN}\\
%\hline
%\astar & 88.0 & 217.3 & 438.1\\
%10-best reranking & 87.7 & - & - \\
%100-best reranking & - & - & - \\
%2-best beam search & 87.6 & 423.5 & 2128.2 \\
%4-best beam search & 87.8 & 1078.7 & 7175.1 \\
%8-best beam search & 87.9 & 2511.1 & 27715.8 \\
%\hline
%\end{tabular}
%\caption{\label{tab:early-update-decoder-comparisons} Comparison of various decoders with the model trained using early-updates.}
%\end{center}
%\end{figure*}
\section{Related Work}
Many structured prediction problems are based around dynamic programs, which are incompatible with recursive neural networks because of their real-valued latent variables. Some recent models have neural factors \cite{durrett:2015}, but these cannot condition on global parse structure, making them less expressive.
Our search explores fewer nodes than dynamic programs, despite an exponentially larger search space, by allowing the recursive neural network to guide the search.

Previous work on structured prediction with recursive or recurrent neural models has used beam search--e.g. in shift reduce parsing~\cite{dyer:2015}, string-to-tree transduction~\cite{vinyals:2014}, or reranking~\cite{socher:2013}--at the cost of potentially recovering suboptimal solutions. 
For our model, beam search is both less efficient and less accurate than optimal \astar decoding. In the non-neural setting, \newcite{zhang:2014} showed that global features with greedy inference can improve dependency parsing.
The CCG beam search parser of~\newcite{clark:2015}, most related to this work, also uses global features. By using neural representations and exact search, we improve over their results.

\astar parsing has been previously proposed for locally factored models~\cite{klein:2003,pauls:2009,auli:2011,lewis:2014b}. We generalize these methods to enable global features. \newcite{vaswani:2016b} apply best-first search to an unlabeled shift-reduce parser. Their use of error states is related to our global model that penalizes local scores. We demonstrated that best-first search is infeasible in our setting, due to the larger search space.%---but combining with a factored model allows an efficient \astar search. 

A close integration of learning and decoding has been shown to be beneficial for structured prediction. \textsc{Searn}~\cite{daume:2009} and \textsc{DAgger}~\cite{ross:2011} learn greedy policies to predict structure by sampling classification examples over actions from single states. We similarly generate classification examples over hyperedges in the agenda, but actions from multiple states compete against each other. Other learning objectives that update parameters based on a beam or agenda of partial structures have also been proposed \cite{collins:2004,daume:2005,huang:2012,andor:2016,wiseman:2016}, but the impact of search errors is unclear.
\section{Conclusion}
We have shown for the first time that a parsing model with global features can be decoded with optimality guarantees. This enables the use of powerful recursive neural networks for parsing without resorting to approximate decoding methods. Experiments show that this approach is effective for CCG parsing, resulting in a new state-of-the-art parser. In future work, we will apply our approach to other structured prediction tasks, where neural networks---and greedy beam search---have become ubiquitous.

%Our \astar parser can efficiently search in a super-exponential space by incorporating existing locally-factored models, for which informative \astar heuristics can be computed. Given the locally-factored model, an additional non-positive global component learns to model the remaining non-local phenomena. 
%We introduced a learning objective that directly optimizes the \astar agenda and encourages the parser to search correctly and efficiently. 

\section*{Acknowledgements}
We thank Luheng He, Julian Michael, and Mark Yatskar for valuable discussion, and the anonymous reviewers for feedback and comments.

This work was supported by the NSF (IIS-1252835,
IIS-1562364), DARPA under the DEFT program
through the AFRL (FA8750-13-2-0019), an Allen
Distinguished Investigator Award, and a gift from
Google.

\bibliography{main}
\bibliographystyle{emnlp2016}

\end{document}